\documentclass[preprint, 12pt, a4paper]{elsarticle}

\usepackage{amssymb}
\usepackage{amsthm}
\usepackage{url}

\usepackage{lineno}

\journal{SoftwareX}

\begin{document}

\begin{frontmatter}



\title{HexagDLy - Processing hexagonally sampled data with CNNs in PyTorch}


\author[potsdam]{Constantin Steppa\corref{cor1}}
\ead{steppa@uni-potsdam.de}
\author[hu]{Tim L. Holch\corref{cor1}}
\ead{holchtim@physik.hu-berlin.de}

\address[potsdam]{Experimental Astroparticle Physics, Department of Physics and Astronomy, University of Potsdam, Karl-Liebknecht-Stra\ss e 24/25, 14476 Potsdam-Golm, Germany}
\address[hu]{Experimental Particle Physics, Department of Physics, Humboldt University of Berlin, Newtonstr. 15, 12489 Berlin, Germany}

\cortext[cor1]{Both authors contributed equally.}

\begin{abstract}

HexagDLy is a Python-library extending the PyTorch deep learning framework with convolution and pooling operations on hexagonal grids. It aims to ease the access to convolutional neural networks for applications that rely on hexagonally sampled data as, for example, commonly found in ground-based astroparticle physics experiments.

\end{abstract}


\end{frontmatter}



\section{Motivation and significance}
\label{sec:intro}

Convolutional neural networks (CNNs) are a powerful and versatile tool in big
data analysis and computer vision~\cite{nature_DL}. Their application has been widely promoted in
various research fields by the availability of open-source deep learning
frameworks (DLFs) like TensorFlow, Caffe, PyTorch or the Microsoft Cognitive
Toolkit. Also in ground-based astroparticle physics experiments, where large
amounts of image-like data need to be analysed, the application of CNNs has come
into focus. 

This data is often hexagonally sampled, which poses an initial
obstacle for the application of CNNs: DLFs cannot process hexagonally sampled
data out-of-the-box. Solutions to this problem have been presented in several
applicability studies \cite{Feng2016, Holch2017, Huennefeld2017,
Erdmann2018, Mangano2018, Shilon2018}. Most of these solutions are based on transforming the hexagonally
sampled data to an approximate representation on a rectangular grid via
pre-processing such as rebinning, interpolation, oversampling and axis-shearing.
HexagDLy, on the other hand, provides a native solution to process hexagonally sampled data.
It relies on a specific addressing scheme for hexagonally sampled data that 
allows for the construction of convolution and pooling operations on hexagonal
grids by using methods provided by PyTorch\footnote{\url{https://pytorch.org/}}~\cite{Paszke2017}.
HexagDLy thereby aims to exploit the benefits of directly processing hexagonally sampled data,
of which the most notable are reduced computing resources \cite{Mersereau1979}, more efficient image processing operators
\cite{Staunton1989} and higher angular resolution \cite{Staunton1990}.

In the context of CNNs, Hoogeboom et al. have already demonstrated the
advantages of applying hexagonal convolutions such as improved accuracies due
to the reduced anisotropy of hexagonal filters \cite{Hoogeboom2018}.
With HexagDLy, hexagonal convolutions are available in an open-source software
with focus on user-friendliness. It facilitates access to CNNs for any
kind of hexagonally sampled data, which, in addition to ground-based
astroparticle physics, can be found in other research fields like
ecology \cite{Birch2007} or numerical climate modeling
\cite{Sahr2011, Satoh2014}.

In the following Sec.~\ref{sec:software} 
the software is described, including its capabilities and the
requirements on the input format. 
The application of HexagDLy is illustrated with an example in
Sec.~\ref{sec:example} followed by a comparative study on the application of hexagonal and square convolution kernels.
Potential benefits of using hexagonal convolutions in
ground-based astroparticle physics are outlined in Sec.~\ref{sec:impact}.

\section{Software description}
\label{sec:software}

HexagDLy provides convolution operations on hexagonal grids built on PyTorch routines. Given the required input format for these routines, an addressing scheme has to be chosen to map the hexagonally sampled data to Cartesian tensors. The convolution and pooling operations are then adapted accordingly to reflect the hexagonal structure of the original data which is also conserved in the output. This is done by constructing custom hexagonal kernels that are applied in combination with a strict padding and striding scheme.
These main ideas behind HexagDLy are outlined below.
Please see Table~\ref{tab_metadata} for the repository and software dependencies.

\subsection{Input Format}
\label{sec:input}

\begin{figure}[h]
    \centering
    \includegraphics[width=\textwidth]{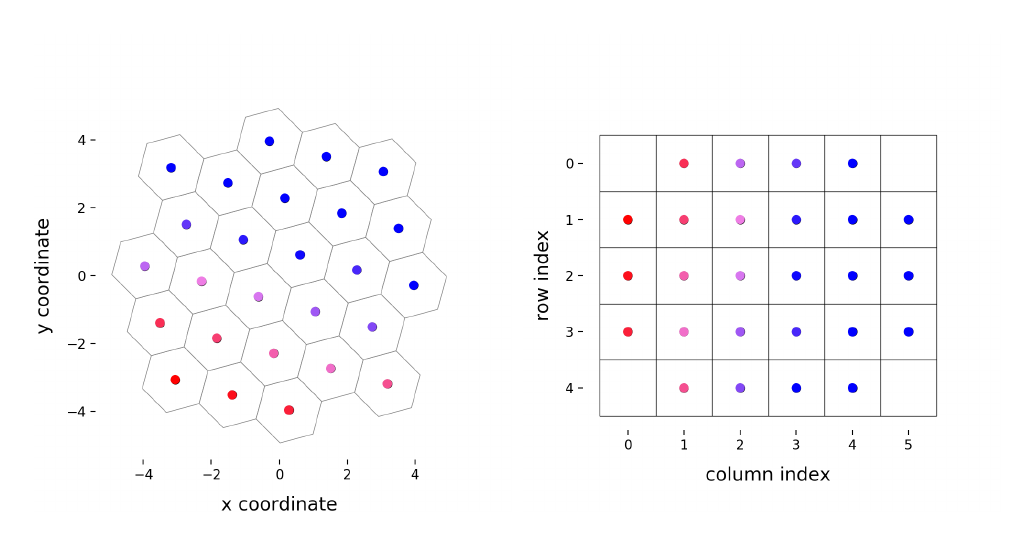}
    \caption{Illustrative example of a hexagonal array with pixel positions
    given in Cartesian coordinates (left) and the corresponding tensor
    indices as inferred from the described addressing scheme (right). 
    Blank elements of the tensor have to be filled with arbitrary values.}
    \label{fig:addressing}
\end{figure}

In order to map hexagonally sampled data to Cartesian tensors, different
addressing schemes can be applied (for example, see \cite{Hoogeboom2018}).
HexagDLy uses the scheme that allows for the most efficient data
storage. As a hexagonal grid can be interpreted as two overlayed rectangular
grids, the data points can be combined in a single square-grid array by aligning
the two rectangular parts. The procedure is illustrated in
Figure~\ref{fig:addressing} where the hexagonal array in Cartesian coordinates is
first rotated to achieve a vertical alignment of neighbouring elements (called
\textit{pixels} hereafter). This allows a separation of the data into columns. The
pixels are then aligned horizontally by shifting every second column upwards by
half the distance between neighbouring pixels, resulting in a square-grid array
with rows and columns. Counting rows from top to bottom and columns from left to
right yields the indices for each element in the input tensor, which corresponds
to a certain pixel in the hexagonal array. Tensor elements that do not have a
corresponding counterpart in the hexagonal array have to be filled with an
arbitrary value.

\subsection{Hexagonal Kernels}
\label{sec:kernels}
 
\begin{figure}[h]
    \centering
    \includegraphics[width=0.5\textwidth]{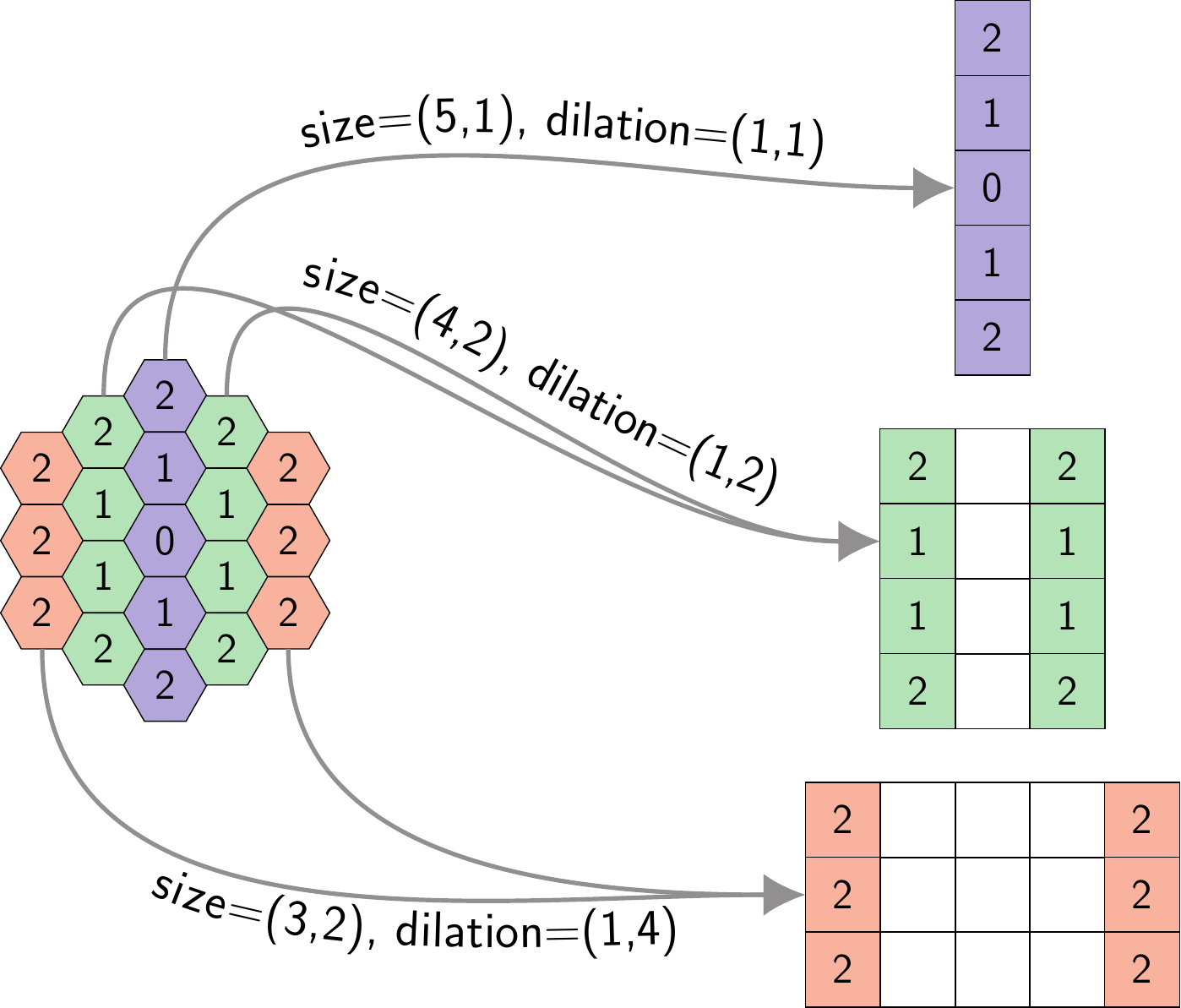}
    \caption{Schematic construction of a hexagonal kernel from rectangular sub-kernels within HexagDLy. 
    Only the exterior columns of sub-kernels contain values while interior columns 
 	are disregarded by setting a \textit{dilation} $\neq 1$ (see the PyTorch documentation for details).}
    \label{fig:kernel}
\end{figure}

The implemented convolution operations use kernels on the hexagonal grid that have a 6-fold rotational symmetry
(i.e. kernels of hexagonal shape). The geometry of a kernel is therefore described only by 
its size which is a single integer corresponding to the number of layers of neighbouring elements 
around its central element. Internally, HexagDLy constructs these hexagonal kernels from rectangular 
sub-kernels as illustrated in Fig.~\ref{fig:kernel}. The illustrated kernel of size 2 consists of 
three sub-kernels, each representing a set of equal-length columns of the hexagonal kernel.
The spatial relation between these columns are accounted for via defined horizontal dilations.

\subsection{Convolution Operations}
\label{}

\begin{figure}[h]
    \centering
    \includegraphics[width=\textwidth]{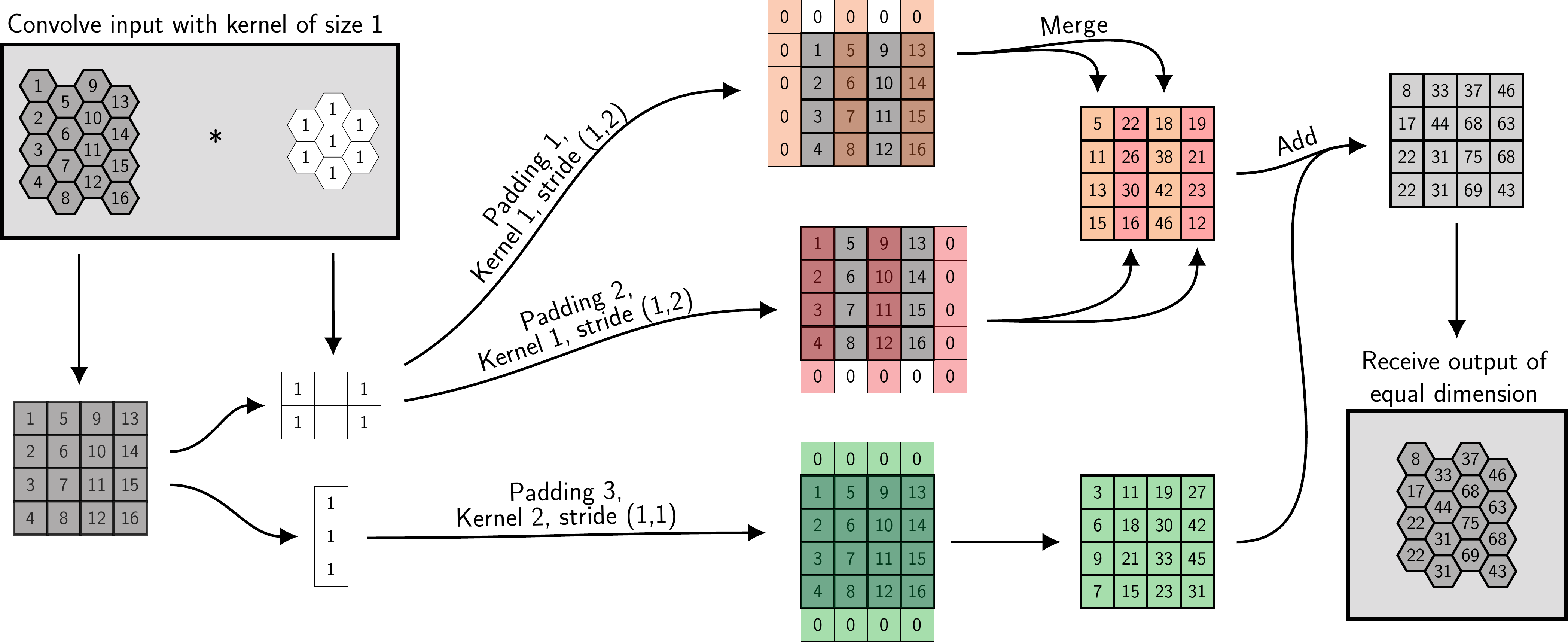}
    \caption{Realisation of a hexagonal convolution with a kernel of size 1 in
    HexagDLy. First, the input data is rearranged into a tensor (as described in
    Sec.~\ref{sec:input}) and the kernel is divided into rectangular
    sub-kernels. For every sub-kernel, different paddings and strides are
    applied to the input to account for the shifted columns. The results of the
    sub-convolutions are then merged and added to receive the convolved hexagonal
    data in tensor format.}
	\label{fig:nn_hex_conv}
\end{figure}

Since a hexagonal kernel is constructed out of multiple rectangular
sub-kernels, a single hexagonal convolution operation is realised by a
combination of multiple convolutions of the input tensor with these sub-kernels. As described in Sec.~\ref{sec:input}, 
columns of the hexagonal array are shifted to match with the tensor format required by PyTorch. The single sub-convolutions therefore have to be adapted in order 
to account for this shift. This is achieved by defining a complex scheme for the
padding and slicing of the input tensor. In this scheme, the number of
rows and columns that are padded or sliced for each sub-convolution depends on the size of
the input tensor as well as on the size of the hexagonal kernel and the applied
stride. To conserve the hexagonal structure of the data, only symmetric strides in equally sized 
steps along the three symmetry axes of the hexagonal grid are performed, starting from the top 
left cell.
Figure~\ref{fig:nn_hex_conv} illustrates the single
steps of this procedure for the convolution of a toy tensor with a hexagonal
kernel of size 1.

\begin{figure}[h]
    \centering
    \includegraphics[width=\textwidth]{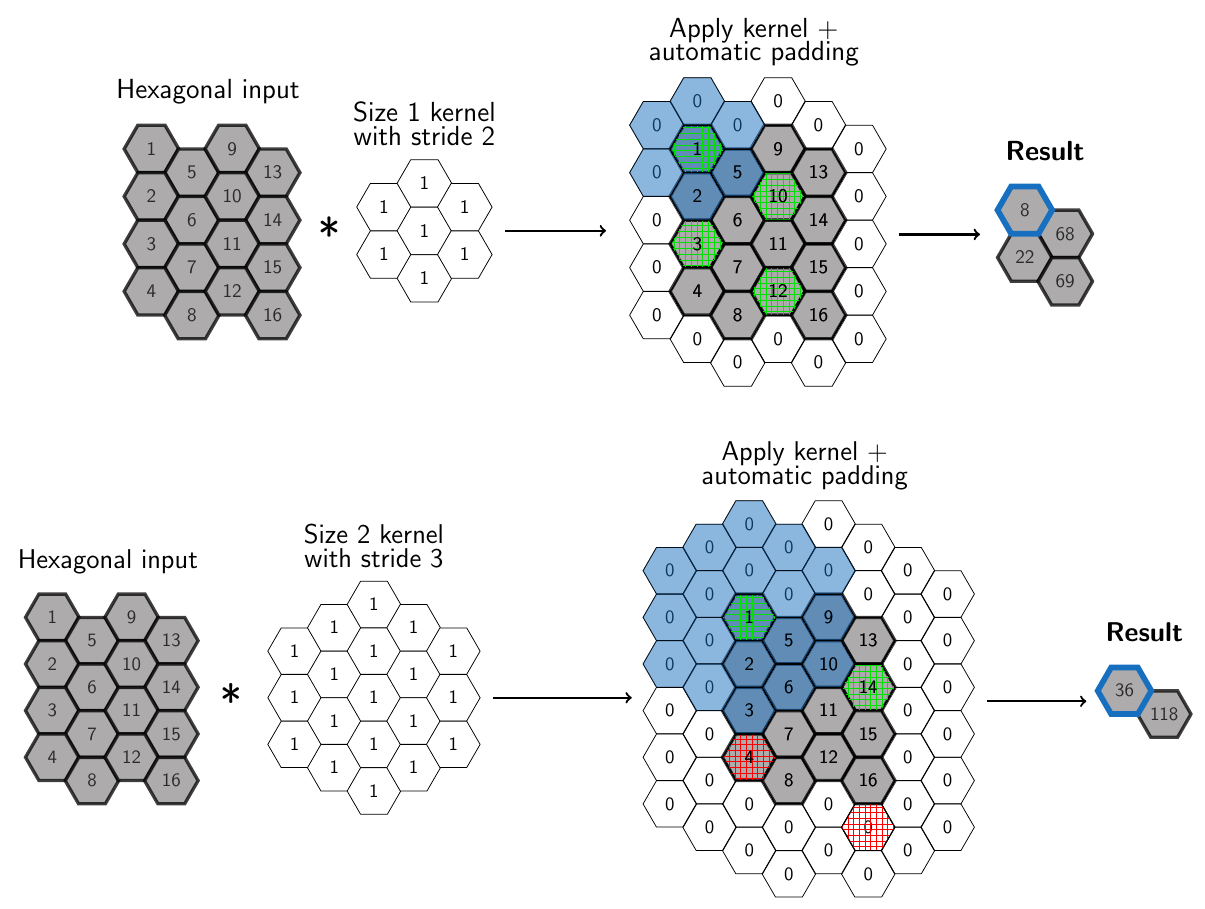}
    \caption{Illustration of the padding and striding scheme for convolutions with
    different kernel sizes and strides. 
    The green and red elements mark valid and omitted steps, respectively.    
    The first position of a kernel is 
    marked in blue as well as its corresponding output cell in the result. 
    }
    \label{fig:padding_strides}
\end{figure}

It is important to note that a kernel is always centred on a pixel that is part
of the actual input tensor and not of the padded rows and columns. To conserve the data format
used by HexagDLy, steps that would lead to an output with 
columns of unequal length are neglected.
Figure~\ref{fig:padding_strides} illustrates this padding and
convolution-element selection for different strides and kernel sizes, including such a case, 
where a step is omitted.

\subsection{Software Functionalities}
\label{}

HexagDLy provides two- and three-dimensional hexagonal convolution operations.
In the three-dimensional case, the input data is expected to have a hexagonal 
layout in the x-y-plane while data points along the z-axis are assumed to be equidistant.
This makes it possible e.g. to process time-resolved data of a
two-dimensional detector with hexagonal layout. Following the design of 
convolution operations, pooling methods are implemented accordingly.
This is done by replacing the PyTorch-based sub-convolutions with the according pooling 
methods and combining the outputs with aggregation functions, whereas the padding- and 
striding-scheme is identical.
By adopting the PyTorch-API, these operations can easily be incorporated in CNN
models defined in PyTorch. Furthermore, it is possible to define custom 
hexagonal kernels with defined values for each kernel element, making
it possible to manually implement structure detecting kernels or to perform data
processing like smoothing on hexagonally sampled data. 
Examples are provided in the online repository in the form of jupyter notebooks (see Tab.~\ref{tab_metadata}) 
that demonstrate the functionalities and usage of the methods provided by HexagDLy.

\section{Illustrative Example}
\label{sec:example}

\begin{figure}[h]
    \centering
    \includegraphics[width=1\textwidth]{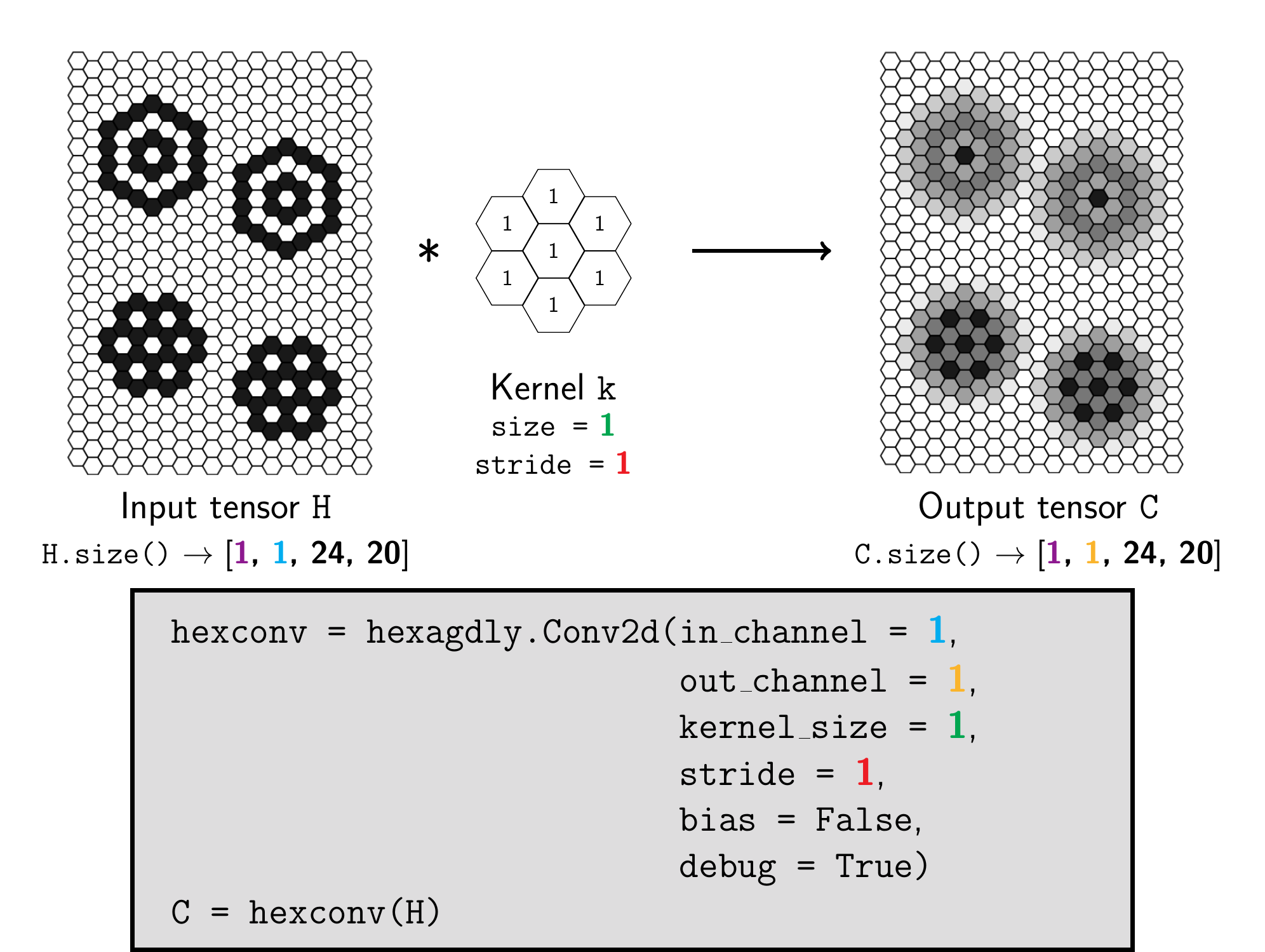}
    \caption{Schematic application of a hexagonal kernel of size 1 to
    hexagonal shapes on a hexagonal grid. The corresponding code is given in the grey box, 
    whereas the parameters defining the operation are colour-coded.
    Enabling the debug mode sets all kernel elements to 1.}
    \label{fig:ie}
\end{figure}

To outline the application of HexagDLy, a set of examples covering basic use-cases 
is provided along with the HexagDLy source code in the online repository. An illustrative way to 
demonstrate the functioning and capabilities of HexagDLy is to perform hexagonal operations 
on hexagonally sampled shapes that themselves exhibit a 6-fold symmetry. 
In Fig.~\ref{fig:ie} the result of convolving an image displaying hexagonal
shapes with a hexagonal kernel is shown. It can clearly be seen that the 6-fold
symmetry of the original shapes on the hexagonal grid is conserved in the output.
For an example of how to use HexagDLy in a CNN, please see the provided jupyter notebooks in the online repository (see Tab.~\ref{tab_metadata}).

\section{Comparing Hexagonal and Square Convolution Kernels}
\label{sec:compare}

As outlined in Sec.~\ref{sec:intro}, a hexagonal sampling of two-dimensional data allows for more efficient data processing compared to a square-grid sampling. Starting with hexagonally sampled data, a conversion to a square grid representation therefore implies less efficient data processing. Additionally, re-sampling hexagonally sampled data to a square grid can introduce sampling artefacts and often requires an increase in resolution to reduce distortions. 

In the context of deep learning, the effects of re-sampling and the reduction of processing efficiency can have a significant influence on the process of designing, optimising and applying CNN-based algorithms.
While the applied re-sampling method is an independent parameter that can be optimised, an increase in resolution demands more computer storage and implies larger convolution kernels or more convolution layers to retain a certain receptive field.
In combination, these effects can influence the performance of a CNN significantly.
This is demonstrated in the following by comparing the performance of CNNs that are trained for the same task but use either hexagonal- or square-grid operations on hexagonal or re-sampled data, respectively. 

\begin{figure}[h]
    \centering
    \includegraphics[width=0.8\textwidth]{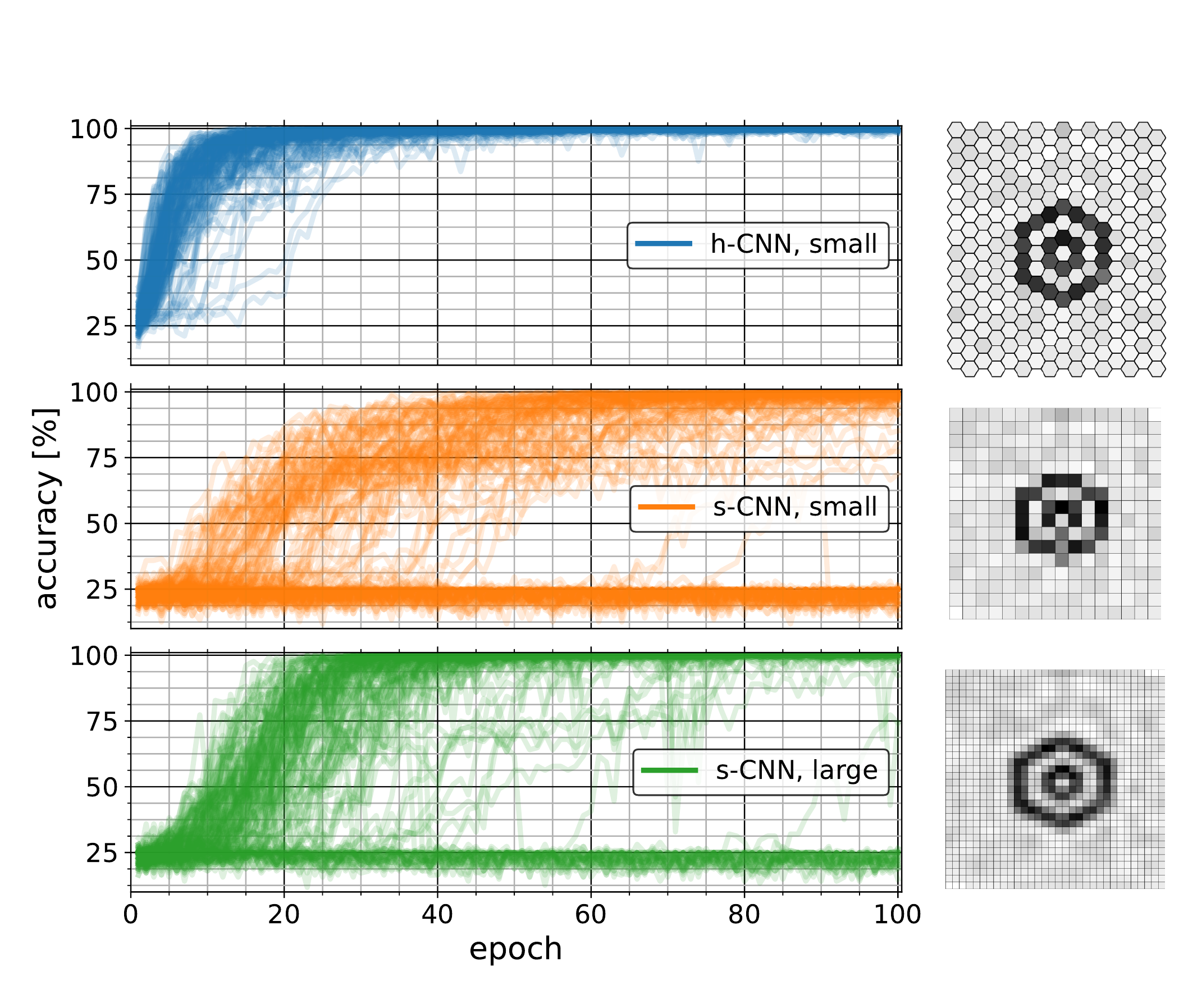}
    \caption{
    Learning curves of 150 iterations for the three CNN models trained to distinguish between four different hexagonal shapes. Example images of one of the four shapes are shown in their different samplings right of each learning curve. See Sec.~\ref{sec:compare} for details on the CNN models and data sets.
 	}
    \label{fig:lcs}
\end{figure}

For the presented experiment, a data set was created with images of four different hexagonal shapes at random positions on a hexagonal grid, overlayed with Gaussian noise. This data set was then interpolated to a square grid of the same resolution (small) as well as to a square grid with four times the number of pixels (large). An example of such a hexagonal shape with the according re-sampled images is shown in Fig.~\ref{fig:lcs}. Two CNN models with the same architecture were set up with the only difference being the use of hexagonal (\textit{h-CNN, small}) or square-grid operations (\textit{s-CNN, small}). These two models have two convolutional and three fully connected layers with a total of $\sim13$k learnable parameters. A third CNN model with three convolutional and three fully connected layers and a total of $\sim1.2$M learnable parameters (\textit{s-CNN, large}) was set up and trained on the large square-grid data. The full implementation of the CNN-models and the data set are provided in a jupyter notebook in the online repository. The three CNNs were trained for $100$ epochs on $128$ images per class with a self-adjusting learning rate. This was repeated $150$ times with the training data being regenerated and the models being reinitialised in each iteration. 

Figure~\ref{fig:lcs} shows the resulting learning curves for all iterations for each CNN-model. It can be seen that the h-CNN reliably reaches $100\%$ accuracy after a few epochs of training. Both s-CNNs, on the other hand, show a generally worse learning behaviour. Although they are both able to achieve $100\%$ accuracy in some cases, only in $60\%$  (small) and $80\%$ (large) of all iterations the models reach accuracies above random guessing performance.

This toy example illustrates the advantages of directly processing hexagonally sampled data in terms of reliability and accuracy.
The differences in performance of the two s-CNNs demonstrate that the effects of re-sampling can be compensated by increasing the resolution of the re-sampled data and likewise extending the CNN capacity. However, even with two orders of magnitude more learnable parameters, the performance of the h-CNN is not reached.
Even though the performance difference between h-CNN and s-CNN may not be as significant in a realistic application, natively processing hexagonally sampled data is generally expected to be the most efficient approach.
However, the current implementation of hexagonal operations in HexagDLy produces a significant computational overhead compared to its according square-grid operation in PyTorch. This can increase the processing time for an h-CNN implemented with HexagDLy, but does not influence the advantages of applying hexagonal convolutions as outlined above.

\section{Impact}
\label{sec:impact}

Hexagonally sampled data is common in ground-based astroparticle physics
experiments like the High Energy Stereoscopic System (H.E.S.S.), the Pierre Auger Observatory or IceCube where large areas have to be 
efficiently covered with a limited number of detectors.
This can be achieved by arranging the detectors on a hexagonal grid as it allows 
for the densest tiling of a two-dimensional Euclidean plane and for optimal sampling of circularly band-limited signals.
In these experiments data is taken at high rates and is 
mostly background-dominated. Additionally, this data can cover a large parameter
space, e.g. multiple telescopes taking data simultaneously.
Therefore, advanced data processing algorithms are used to analyse this data. The application of machine
learning techniques has already become a standard in this respect
\cite{Ohm2009, Aartsen2015}. Following the progress in the field of machine learning, CNNs 
represent promising means to further improve data analyses for astroparticle physics experiments.

By providing convolution and pooling operations that
can be directly applied to hexagonally sampled data, HexagDLy provides a
user-friendly environment to explore the applicability of CNNs for these experiments.
Since no pre-processing is required, the initial efforts for the application of CNNs can be
significantly reduced compared to other approaches. 

The increasing scales and sensitivity of future
observatories like the Cherenkov Telescope Array \cite{CTA2017} will result in much larger data sets that
need to be analysed. This will pose additional challenges for the
analysis in terms of performance and resources. The methods provided by HexagDLy
can help to address these challenges.

\section{Conclusions}
\label{}

Following the growing interest in CNNs, increasing efforts to adapt convolution operations to non-Cartesian data can be observed, as for example for spherical data \cite{Taco2018, Perraudin2018} and non-Euclidean manifolds \cite{Masci2015}. Besides \cite{Hoogeboom2018}, HexagDLy presents a solution for hexagonally sampled data.
With a focus on flexibility and user-friendliness, HexagDLy provides convolution and pooling operations on
hexagonal grids. It is based on PyTorch and makes use of the torch.nn module for the implementation of these operations. 
In combination with a special data addressing scheme, it facilitates the access to CNNs for hexagonally sampled data.
By taking advantage of the benefits of directly processing
hexagonally sampled data, HexagDLy aims to promote research based on the applicability of CNNs 
e.g. in ground-based astroparticle physics.
Currently, HexagDLy is used in a study on the
applicability of CNNs for the analysis of data from the H.E.S.S. experiment. A report on first results is in preparation.

\section*{Acknowledgements}
\label{}

This software project evolved as part of a research study to explore deep learning algorithms in the analysis of imaging Cherenkov telescope data within the H.E.S.S. collaboration. We acknowledge the support of the whole collaboration in this project. 
In particular we want to thank our colleagues Matthias B\"uchele, Kathrin Egberts, Tobias Fischer, Manuel Kraus, Thomas Lohse, Ullrich Schwanke, Idan Shilon and Gerrit Spengler for fruitful discussions that promoted the development of HexagDLy.



\bibliographystyle{elsarticle-num} 
\bibliography{hexagdly_paper}


\section*{Current code version}
\label{}


\begin{table}[!h]
\begin{tabular}{|l|p{6.5cm}|p{6.5cm}|}
\hline
\textbf{Nr.} & \textbf{Code metadata description} & \textbf{Please fill in this column} \\
\hline
C1 & Current code version & 2.0.1 \\
\hline
C2 & Permanent link to code/repository used for this code version & \url{https://github.com/ai4iacts/hexagdly/tree/2.0.1}\\
\hline
C3 & Legal Code License   & MIT\\
\hline
C4 & Code versioning system used & git\\
\hline
C5 & Software code languages, tools, and services used & Python~3, PyTorch \\
\hline
C6 & Compilation requirements, operating environments \& dependencies & Tested on Linux and Mac OS, Python $\geq$ 3.6, PyTorch $\geq$ 0.4\\
\hline
C7 & If available Link to developer documentation/manual & \url{https://github.com/ai4iacts/hexagdly/README.md} \\
\hline
C8 & Support email for questions & steppa@uni-potsdam.de, holchtim@physik.hu-berlin.de\\
\hline
\end{tabular}
\caption{Code metadata (mandatory)}
\label{tab_metadata} 
\end{table}

%
%

\end{document}